\documentclass{article}

\PassOptionsToPackage{numbers, compress}{natbib}

\usepackage[final]{neurips_2022}

\usepackage[utf8]{inputenc} 
\usepackage[T1]{fontenc}    
\usepackage{hyperref}       
\usepackage{url}            
\usepackage{booktabs}       
\usepackage{amsfonts}       
\usepackage{nicefrac}       
\usepackage{microtype}      
\usepackage{xcolor}         
\usepackage{amsmath}
\usepackage{amssymb}
\usepackage{graphicx}
\usepackage{caption}
\usepackage{subcaption}
\usepackage{multirow}
\usepackage{amsthm}
\usepackage{comment}

\title{Xtal2DoS: Attention-based Crystal to Sequence Learning for Density of States Prediction}

\author{%
  Junwen Bai\thanks{Work done at Cornell University.} \\
  Google \\
  \texttt{junwen@google.com} \\
  \And
  Yuanqi Du \\
  Cornell University \\
  \texttt{yd392@cornell.edu} \\
  \And
  Yingheng Wang \\
  Cornell University \\
  \texttt{yw2349@cornell.edu} \\
  \And
  Shufeng Kong \\
  Cornell University \\
  \texttt{sk2299@cornell.edu} \\
  \And
  John Gregoire \\
  Caltech \\
  \texttt{gregoire@caltech.edu} \\
  \And
  Carla Gomes \\
  Cornell University \\
  \texttt{gomes@cs.cornell.edu} \\
}

\begin{document}

\maketitle

\begin{abstract}
Modern machine learning techniques have been extensively applied to  materials science, especially for property prediction tasks. A majority of these methods address  scalar property predictions, while more challenging spectral properties remain less emphasized. We formulate a crystal-to-sequence learning task and propose a novel attention-based learning method, Xtal2DoS, which decodes the sequential representation of the material density of states (DoS) properties by incorporating the learned atomic embeddings through attention networks. Experiments show Xtal2DoS is faster than the existing models, and consistently outperforms other state-of-the-art methods on four metrics for two fundamental spectral properties, phonon and electronic DoS.
\end{abstract}

\section{Introduction}

Machine learning for materials discovery has risen to prominence recently, with impressive applications from materials design \cite{sanchez2018inverse,du2022molgensurvey,du2022chemspace} to materials' property predictions \cite{liu2017materials,schmidt2019recent}. While the performance greatly exceeds the existing models, most of the achievements are confined to individual scalar quantities \cite{xie2018crystal, isayev2017universal, de2016statistical,zhuo2018predicting}. For conventional machine learning models, some other structured properties are often hard to deal with, for example spectral properties  \cite{eismann2012spectral}. Two fundamental spectral properties are phonon density of states (phDoS) \cite{osborn2001phonon} and electronic density of states (eDoS) \cite{heremans2008enhancement}. In a given frequency interval, phDoS describes the number of phonon modes if the density of wave vectors in the Brillouin zone is homogeneously distributed. The phonon density of states is typically normalized to one \cite{parlinski2010computing}. Similarly, eDoS depicts the number of different states at a particular energy level that electrons are allowed to occupy (namely, the number of electron states per unit volume per unit energy) \cite{omar1975elementary}. Mathematically, the density of states can be regarded as a sequence depicting the fundamental characteristics of particles. Each crystalline material is associated with unique density of states (phDoS, eDoS). The goal of density of states prediction is to estimate this sequence given an input structure.

There are two challenges in this sequence prediction problem. First, the input crystals are 3D structures, which are very different from typical machine learning formats (i.e., vectors or matrices). Featurizing the crystal structures is the cornerstone of the whole learning process. If useful information cannot be extracted effectively, the follow-up predictions would be severely hampered. Essentially, each crystal structure is a collection of atoms. Each atomic site contains a specific element, and these sites can be seen as nodes on a graph. Atoms also have interactions with each other \cite{martin2020electronic}. Close atoms often have strong bonds. These bonds can be represented by the connecting edges on the graph \cite{xie2018crystal}. Conventional CNN \cite{krizhevsky2012imagenet} and RNN \cite{hochreiter1997long} are not applicable to graphs, while graph neural networks \cite{hamilton2017inductive} are specifically designed for such structures. The pioneering work applying GNN networks to materials property prediction was CGCNN \cite{xie2018crystal}, where the crystal unit cell was formalized as a graph with atoms as nodes and bonds as edges. MEGNet \cite{chen2019graph} further introduced a global state to add extra features like temperature, pressure, and entropy. More recently, GATGNN \cite{louis2020graph} suggested using an attention mechanism to capture the global crystal structure beyond the local atomic environments. In GATGNN, the local augmented graph-attention layers encode the properties of local environments for each atom to obtain an abstract feature, and a global attention layer aggregates these atomic vectors by weighted averaging to produce the global feature vector for prediction. This exploitation of structure markedly facilitated feature extraction from the crystalline systems, and built an extensible foundation for various predictions.

The second challenge in sequence prediction is the target format. Works like GATGNN focused on the prediction of scalar properties such as formation energies, band gaps and elastic moduli. The performance of these models degrades when applied to spectral properties like eDoS and phDoS \cite{kong2022density}. The simplest way to adapt the previous methods to predict sequences is to expand the last layer to output multiple values instead of a single value. The general idea is similar to predicting multiple scalar properties simultaneously. However, such parallel prediction overlooks the sequential nature of the target DoS. The smoothness property cannot be enforced onto the predicted targets either. Mat2Spec \cite{kong2022density} observes this flaw and re-formulates the problem as a multi-target regression task. Mat2Spec keeps the framework of expanding the last layer to output more values, but additionally uses contrastive learning \cite{henaff2020data} and a probabilistic variational autoencoder \cite{doersch2016tutorial} to model the correlations among different locations of the sequence.

In this work, we introduce Crystal-to-DoS (Xtal2DoS) learning, to improve both the graph encoding and the sequence decoding for phDoS and eDoS predictions. First, we inherit the general framework of GATGNN and employ graph attention networks (GAT) \cite{velivckovic2017graph} to learn both the local and global attention. 
Second, we adopt the attention-based transformer to decode the target sequence \cite{vaswani2017attention}. The transformer reads the one-hot position embedding together with the global crystal representation, and decodes left-to-right auto-regressively. Differing from the multi-target regression perspective, Xtal2DoS takes the sequence-to-sequence (seq2seq) learning perspective. In seq2seq, the decoding process incorporates both  self-attention and  source-attention. Self-attention retrieves the historical memory from the previous steps, and source-attention derives the similarity between the decoding sequence and the input sequence. In this DoS prediction task, the input is a graph rather than a sequence, so the source-attention attends each position with all the atom embeddings. The resulting model, Xtal2DoS, thus improves both the encoding and decoding over prior models, for this graph-to-sequence learning problem. Experimental results show our method can outperform the existing state-of-the-art models on all four evaluation metrics by over 14\% on average for phDoS, and over 4\% on average for eDoS.

\begin{figure}[t]
  \centering
  \includegraphics[width=0.5\linewidth]{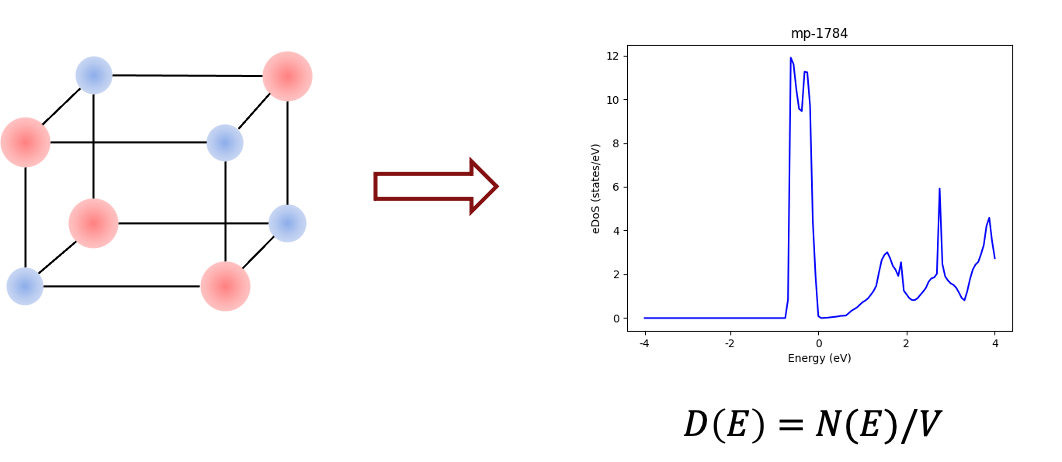}
  \caption[DoS prediction]{An illustration of predicting eDoS from the crystal structure for material \textit{CsF}.}
  \label{fig:dos_pred}
\end{figure}

\section{Crystal-to-DoS Learning}

\subsection{Problem Formulation}

The input crystal material $x$ is denoted as a graph $G_x=(V, E)$. $V=\{u_i\}_{i=1}^n$ is the collection of $n$ atoms where $u_i$ is an atom embedding learnt from CGCNN. $E=\{e_{ij}\}_{1\le i,j\le n}$ denotes the bonds connecting atoms. $e_{ij}$ depicts the distance between two atoms $i, j$. For a crystalline system, the crystal structure repeats in the 3D space and each atom has a 3D coordinate. Though theoretically, each atom has connections with any other atom, for simplicity and generality, we choose $n_{max\_nbr}$ neighbors to be the upper bound for the number of neighbors of each atom. In other words, the sum of each row or column of $G_x$'s adjacency matrix is smaller than or equal to $n_{max\_nbr}$. Our goal is to predict $y$ for each $x$. $y$ is the desired density of states (phDoS or eDoS), with length $l_y$. $l_y=51$ for phDoS and $l_y=128$ for eDoS. $l$ is predefined for each type of DoS. The learnt model maps $x$ to $y$ ($f:x\to y$). Figure \ref{fig:dos_pred} presents an example (caesium fluoride
, \textit{CsF}) of DoS prediction.

\subsection{Graph Attention Encoder}
\label{subsubsec:gat_enc}

As the prototype of GAT, graph convolutional network (GCN) \cite{kipf2016semi} combines the local graph structure and node-level features to yield impressive node predictions. However, the feature aggregation of GCN depends on structure and is mostly linear, constraining effectiveness of message passing on the graph. In GCN, a graph convolution operation can be written as
\begin{equation}
    h_{i}^{l+1}=\sigma\left(\sum_{j\in N(i)}\frac{1}{c_{ij}}W^{l}h_j^{l}\right)
\label{eq:gcn}
\end{equation}
$\sigma(\cdot)$ is a non-linear activation function. $W_l$ is the feature transformation matrix in the $l$-th layer. $h_i^l$ is the latent node embedding for node $i$ in the $l$-th layer, with $h_i^0=u_i$. $c_{ij}=\sqrt{|N(i)||N(j)|}$ is a normalization constant and $N(i)$ is the set of node $i$'s neighbors. 

In Xtal2DoS, we use a variant of GAT, \textit{UniMP} \cite{shi2020masked}, to encode the crystalline graph. Compared to GAT, UniMP adds support for edge features. For each layer in UniMP, $h_i^l$ is first linearly transformed to queries, keys and values 
\begin{equation}
\begin{aligned}
    q_i^l = W_q^l h_i^l + b_q^l,~~~~~~
    k_i^l = W_k^l h_i^l + b_k^l,~~~~~~
    v_i^l = W_v^l h_i^l + b_v^l.
\end{aligned}
\end{equation}
Similarly, we also transform the edge features between node $i$ and node $j$
\begin{equation}
\begin{aligned}
    g_{ij} = W_e e_{ij} + b_e.\\
\end{aligned}
\end{equation}
The edge feature transformation is not layer-dependent and $W_e, b_e$ are shared for all edges. The attention coefficients are then computed as follows
\begin{equation}
\begin{aligned}
    \alpha_{ij}^l = \frac{\langle q_i^l,~k_j^l+g_{ij} \rangle}{\sum_{p\in N(i)}\langle q_i^l,~k_p^l+g_{ip} \rangle}
\end{aligned}
\end{equation}
where $\langle q,k \rangle=\exp(\frac{q^Tk}{\sqrt{d}})$ and $d$ is the dimensionality of $q_i^l, k_i^l$ or $v_i^l$. 
Afterwards, the embeddings from neighbors are aggregated, together with the edge embeddings, weighted by attention scores
\begin{equation}
    \hat{h}_{i}^{l+1}=\sum_{j\in N(i)}\alpha_{ij}^l\cdot (v_j^l + e_{ij}).
\label{eq:gat}
\end{equation}
From Eq. \ref{eq:gcn} and Eq. \ref{eq:gat}, GAT considers similarity scores and edge information when performing the aggregation to gain more expressiveness. 

Finally, motivated by the success of residual neural networks \cite{he2016deep}, UniMP appends a gated residual connection before passing $\hat{h}_{i}^{l+1}$ to the next layer
\begin{equation}
\begin{aligned}
    r_i^l &= W_r^l ~ h_i^l + b_r^l, \\
    \beta_i^l &= \text{sigmoid}(a^l ~[\hat{h}_{i}^{l+1} ~;~ r_i^l ~;~ \hat{h}_{i}^{l+1}-r_i^l])\\
    h_{i}^{l+1}&=\sigma(f_{LN}((1-\beta_i^l)~\hat{h}_{i}^{l+1}+\beta_i^l~r_i^l))
\end{aligned}
\end{equation}
where $;$ denotes concatenation, $a^l$ is the learnable weight parameter, and $f_{LN}$ represents the layer norm. Note that for the last layer of UniMP, $\sigma(\cdot)$ and $f_{LN}$ are skipped.

\subsection{Sequential Decoding}

GAT learns embeddings for each node, $h_i^L$, if there are $L$ layers in total. GATGNN and Mat2Spec simply take the average of all the embeddings as the global feature vector
\begin{equation}
h_x=\frac{1}{n}\sum_{i=1}^n h_i^L
\end{equation}
Then the global feature vector is used to decode the target DoS, using MLP or VAE.

We compared various sequence models including RNN and transformer for decoding. By default, we used the GRU model for RNN.

\begin{figure}[t]
\centering
  \begin{subfigure}{0.75\textwidth}
  \centering
  \includegraphics[width=\textwidth]{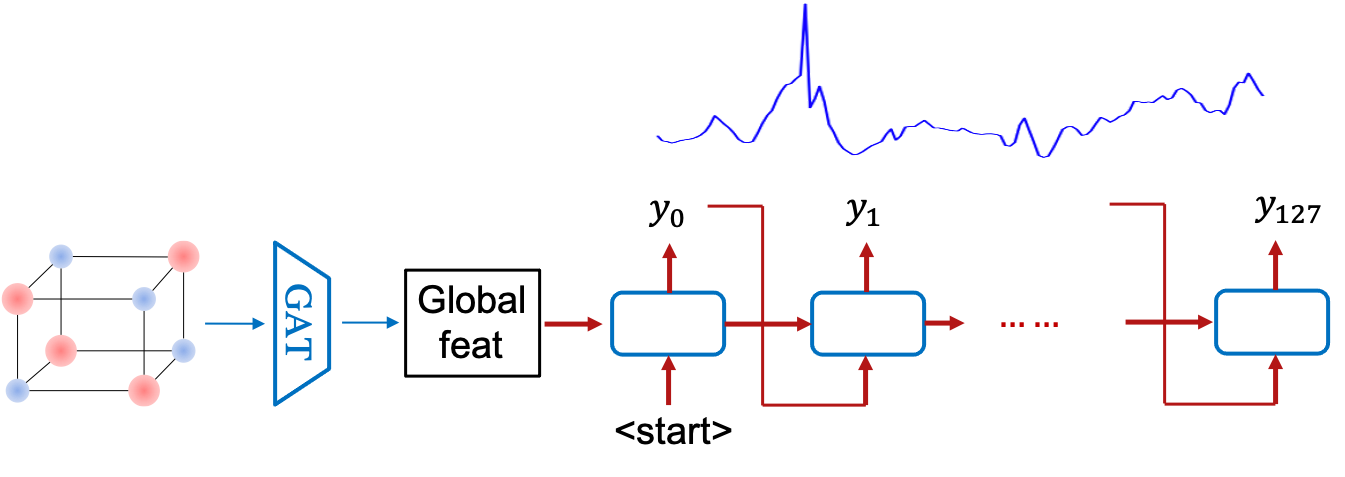}
  \caption[RNN decoding for DoS prediction]{The global feature vector is fed into RNN as the initial state. The decoding unrolls step by step.}
  \label{fig:dos_rnn}
  \end{subfigure}
  ~~
  \begin{subfigure}{0.75\textwidth}
  \centering
  \includegraphics[width=\textwidth]{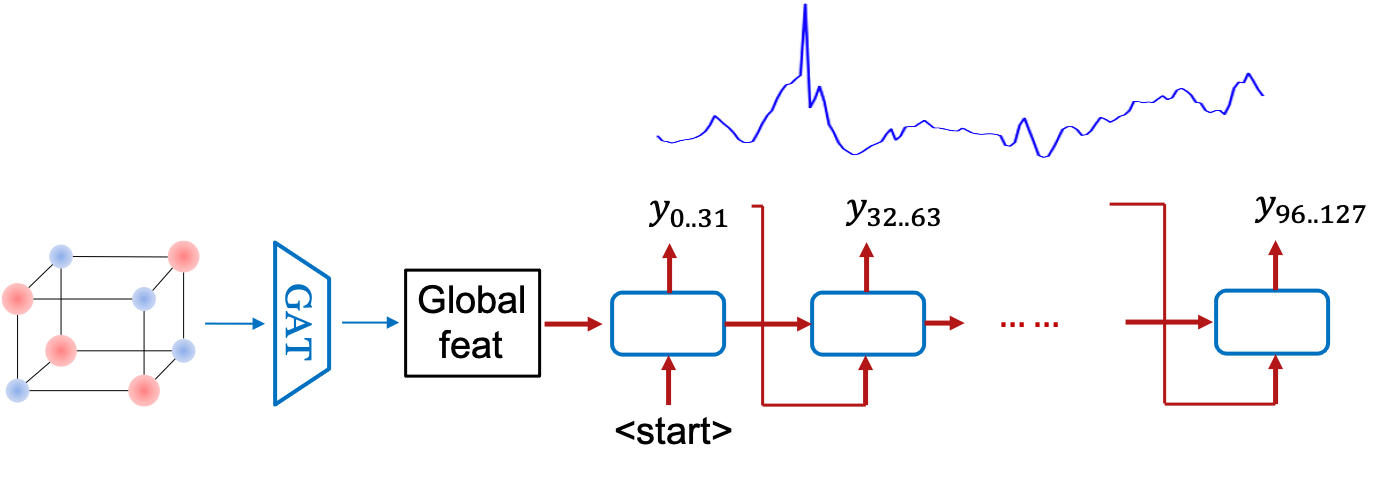}
  \caption[Chunk RNN decoding for DoS prediction]{Chunk RNN decoder is an RNN decoder that predicts a segment instead of a single value at each step.}
  \label{fig:dos_chunk_rnn}
  \end{subfigure}
  \caption{Vanilla RNN, chunk RNN (without attentions).}
\end{figure}

\noindent\textbf{RNN~~} Figure \ref{fig:dos_rnn} sketches the RNN decoding of the GAT-encoded crystal. RNN is naturally suitable for sequence decoding. It reads the global feature vector as the cell state and a special $\langle start\rangle$ symbol to indicate the start of the decoding. The output at each step is also the input for the next step. However, RNN is criticized for lacking long-term memory \cite{zhao2020rnn}. When we decode the last step ($y_{127}$), the memory of the first step ($y_0$) is probably degraded. This might hurt the prediction quality for long sequences. Another drawback of RNN is the speed. Each step has to wait for the previous output, and thus the decoding process could take a very long time.

\noindent\textbf{Chunk RNN~~} Figure \ref{fig:dos_chunk_rnn} gives an overview of a chunk RNN, which is a simple RNN variant. Chunk RNN predicts a segment rather than a single value at each step. The benefit is that the decoding time could be greatly shortened. For instance, on eDoS, if the segment length is 32, then sequence length can be reduced from 128 to 4. On the other hand, each segment still waits for the generation of the previous segment, and longer segments harm the accuracy gradually.

\begin{figure}[t]
\centering
  \begin{minipage}{0.75\textwidth}
  \centering
  \includegraphics[width=\textwidth]{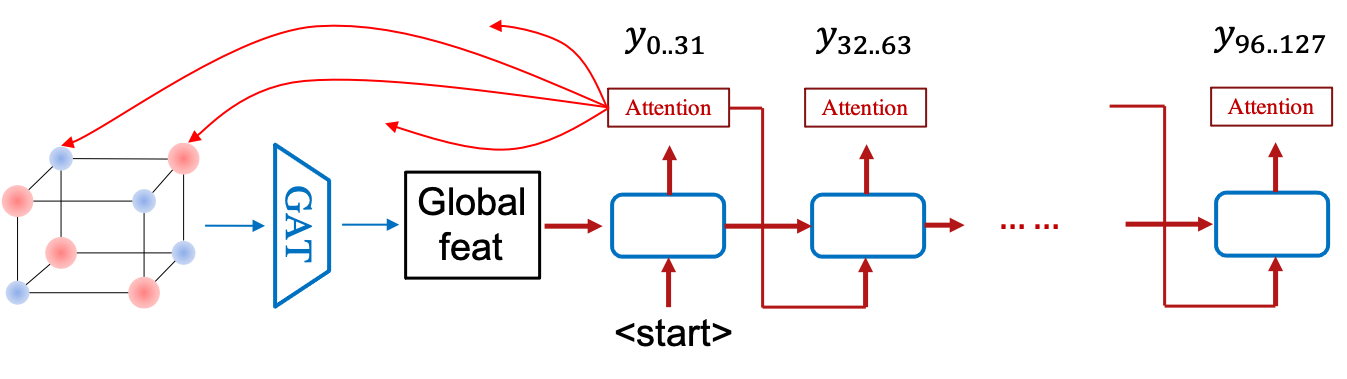}
  \caption{Adding an attention mechanism to the chunk RNN decoder.}
  \label{fig:dos_chunk_rnn_attn}
  \end{minipage}
  \begin{minipage}{0.75\textwidth}
  \centering
  \includegraphics[width=\textwidth]{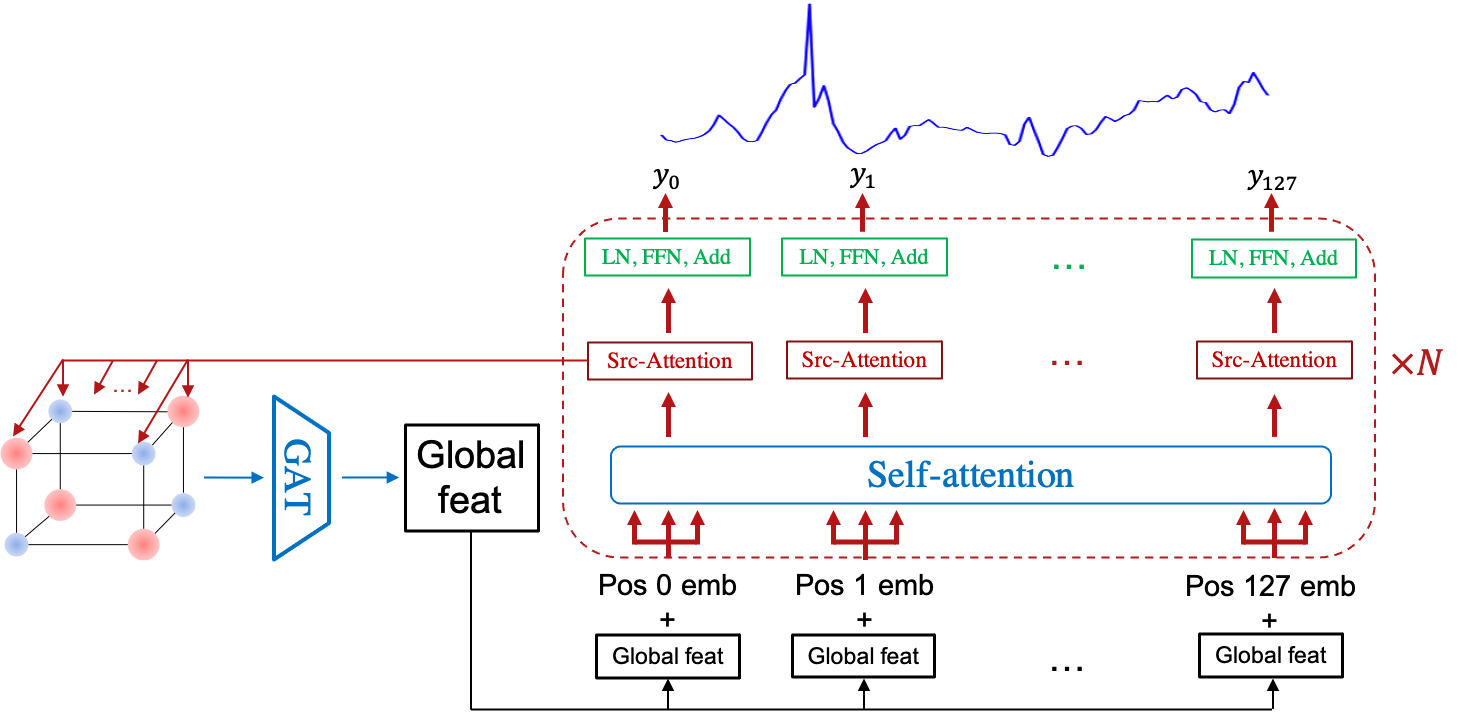}
  \caption{Fully attention-based transformer decoding for DoS prediction. All the recurrent components are replaced with self-attention.}
  \label{fig:dos_transformer}
  \end{minipage}
\end{figure}

\noindent\textbf{Chunk RNN + Attention~~} To compensate for the performance degradation brought by chunk RNN, we further introduce the attention mechanism into the sequence model (Figure \ref{fig:dos_chunk_rnn_attn}). The attention module directly attends to the latent sequence states with the atom embeddings. The latent states can take specific atoms into consideration and build more fine-grained correlations. In practice, we also find including the attention module boosts the evaluation results.

\noindent\textbf{Transformer~~} Given that we have already introduced the attention mechanism, it is intuitive to completely drop the recurrent framework and embrace the fully attentional transformer \cite{vaswani2017attention}. The input at each step is the concatenation of the positional embeddings and the global feature vector. A positional embedding is simply a one-hot vector of length $l_y$ indicating the index of the current step. The global feature vector $h_x$ additionally provides the global context for decoding. Formally, let $z_i=[s_i~;~h_x]$ where $s_i$ is the one-hot position embedding with length $l_y$. $z_i$ is the initial state for the first self-attention layer. As in UniMP, we derive collections of queries, keys and values for the hidden states of each self-attention layer, denoted by Q, K, and V, respectively.

Transformer adopts dot-product attention to calculate the outputs
\begin{equation}
    \text{Attention}(Q,K,V)=\text{softmax}(QK^T/\sqrt{d})~V
\end{equation}
where $d$ is the latent dimensionality. Multiple sets of Q, K, and V matrices can be learned, and each set is known as a head. Multiple heads (multi-head) can give self-attention greater power of discrimination. The outputs of all the heads are concatenated, layer-normalized and decoded through a fully connected feed-forward network. A more detailed depiction can be found in \cite{vaswani2017attention}.

During the decoding phase, each attention layer also computes the source-attention  between the output states and the input atom nodes. If we denote the queries, keys, values in decoding as $Q_{dec}, K_{dec}, V_{dec}$, and the ones from the input graph encoding as $Q_{enc}, K_{dec}, V_{dec}$, we have
\begin{equation}
\begin{aligned}
\text{Self-Attention}(Q_{dec}, K_{dec}, V_{dec})&=\text{softmax}(\frac{Q_{dec}K_{dec}^T}{\sqrt{d}})V_{dec},\\
\text{Source-Attention}(Q_{dec},K_{enc},V_{enc})&=\text{softmax}(\frac{Q_{dec}K_{enc}^T}{\sqrt{d}})V_{enc},
\end{aligned}
\end{equation}
for the sequential decoding. Source-attention usually happens after self-attention. Furthermore, the transformer model adds masking to prevent the current positions from attending to the future positions, to prevent leftward information flow in the decoder, and to preserve the auto-regressive property, ensuring each decoded state only knows the history but not the future. The final generated length-$l_y$ sequence $\hat{y}$ is compared with the ground-truth $y$ through Kullback–Leibler divergence (KL-div) \cite{kullback1997information} or its generalized version, \textit{Bregman Divergence} (BD) \cite{bregman1967relaxation}, resulting in the loss term, $\mathcal L=D_{KL}(y ~||~ \hat{y})$. We chose KL-div and BD because they are more suitable than other measures for data containing peaks, which are ubiquitous in DoS.

Our model is a synergistic combination of a GAT (UniMP) encoder and a transformer decoder. The whole model is entirely based on attentions on graphs and sequences.
Figure \ref{fig:dos_transformer} illustrates the main idea of Xtal2DoS. 

\section{Related Works}

Learning for 3D structures has been a long-standing problem. In materials discovery, two atomic systems can be exactly the same even though the coordinates are totally different in the 3D space if they can be transformed to one another through translation, rotation or inversion. For standard machine learning models, it is very hard to capture such invariance and symmetry \cite{cohen2016group}. There are in general three strategies to address the concern.

First, the same models can be trained using more data \cite{shorten2019survey}. The structured inputs can be randomly transformed without changing the labels. Some prior works successfully applied data augmentation
to exploring materials design spaces \cite{kim2021deep}. On the other hand, in materials science, even with augmentation, it is not easy for machine learning models to comprehend the crystalline system, let alone the data collection and augmentation require enormous domain knowledge and expert effort \cite{schmidt2019recent}.

Second, rather than expanding the scale on the dataset side, some researchers have explored the possibility of enforcing the invariance or equivariance inside the model. This kind of model handles the coordinates directly and are often called equivariant neural networks \cite{satorras2021n,du2022se}. One notable model for materials science applications is E3NN \cite{chen2021direct}. E3NN performs training based on radial functions and spherical harmonics \cite{esteves2018learning}. Intuitively, E3NN represents the atoms with the 3D coordinates, mass and velocity, and tells the equivariant convolutional filters how the crystals might transform. Thus, E3NN can capture the full crystal symmetry, while reducing training set sample complexity. However, radial functions and spherical harmonics may limit the expressiveness of the model. In some scenarios where the data is abundant, E3NN's benefits might become marginal.

Third, it is often helpful to pre-process the data for simplicity \cite{kumar2018introduction}. Instead of using the coordinates, we can re-format the crystals to graphs by retaining only the atoms and the bonds between them. Some of the structure information is lost from the input, such as the lattice angles, but this could also reduce the risk of overfitting. Once the inputs are converted to graphs, mature graph neural network techniques \cite{wu2020comprehensive, huang2022going} can be applied. A representative work is GATGNN \cite{louis2020graph}. GATGNN is one of the primary methods to encode the atom embeddings and extract the global feature vector. It still leads in the performance of many scalar property prediction tasks including absolute energy, Fermi energy, band gap and Poisson-ratio. Nevertheless, in sequence property predictions, its performance degrades significantly \cite{kong2022density}. Mat2Spec \cite{kong2022density} uses the same graph encoder as GATGNN, but improves the decoder. GATGNN only uses an MLP, which is incapable of sequence decoding. Mat2Spec not only aligns the inputs and targets in a latent multivariate Gaussian space, but also learns an embedding for each energy level through contrastive learning. Mat2Spec views the DoS prediction as a multi-target regression problem, yet ignores the sequential structure of DoS. Besides, Mat2Spec simply deepens the MLP decoder and plugs contrastive modules into the decoder, while sequence modeling is cleaner and simpler. 

Xtal2DoS follows the design of GATGNN and Mat2Spec because the graph encoding is shown to outperform the coordinate encoding \cite{kong2022density}. On top of the existing designs, Xtal2DoS further replaces GAT encoder with a more advanced variant, UniMP, and replaces the decoder with a powerful sequence model, transformer. Evaluation results and ablation studies prove the effectiveness and efficiency of our introduced model.

\section{Experiments on DoS Prediction}

We validate the performance of Xtal2DoS on two datasets, phDoS and eDoS. The task of directly predicting phDoS in crystalline solids from atomic species and positions was first proposed in E3NN~\footnote{https://github.com/zhantaochen/phonondos\_e3nn/blob/main/phdos\_train.ipynb} \cite{chen2021direct}. We also include a new version of E3NN, denoted by E3NN-new~\footnote{https://docs.e3nn.org/en/stable/api/nn/models/gate\_points\_2101.html} which improves over the previous version. phDoS is normalized to 1 since phDoS gives the number of modes per unit frequency per unit volume of real space and is essentially a distribution. phDoS is critical for calculating properties such as the average phonon frequency and the heat capacity at 300K. The analogous task of predicting eDoS for nonmagnetic materials was proposed in \cite{kong2022density}. eDoS is the same as phDoS except that it depicts electrons. Materials scientists are interested in a small energy range between -4 to 4 eV with respect to band edges with 63 meV intervals, including both unoccupied and occupied states. Fermi energy and band edges are all set to 0 eV on this energy grid. Predicting eDoS accurately can be very valuable to bandgap energy estimation. Both phDoS and eDoS are continuous functions in nature. We sample 51 points on phDoS and 128 points on eDoS spanning the frequency or energy range. In other words, $l_y=51$ for phDoS and $l_y=128$ for eDoS.

\subsection{Evaluation Metrics}

Four metrics are adopted for evaluation: coefficient of determination ($R^2$), mean absolute error (MAE), mean squared error (MSE) and Wasserstein distance (WD). 

$R^2$ measures the proportion of the variation in the ground-truth targets $y$ that is predictable from the predictions $\hat{y}$, $R^2=1-\frac{\sum_{i=1}^{l_y}(y_i-\hat{y}_i)^2}{\sum_{i=1}^{l_y}(y_i-\bar{y})^2}$,
where $\bar{y}=\frac{1}{l_y}\sum_i y_i$. In the best case when $y$ and $\hat{y}$ match exactly, $R^2$ would reach the maximum possible value, 1. If $y_i=\bar{y}$, $R^2=0$. Note that $R^2$ score can be negative if the predictions are worse. 

MAE and MSE are standard metrics for regression. We also adopt them for  evaluation.

Wasserstein distance is another metric defined between probability distributions. If we view the two distributions as two piles of earth, WD is the minimum amount of dirt to move in order to turn one pile to be the same as the other. Therefore, WD is also known as earth mover's distance. We define the WD as
\begin{equation}
\begin{aligned}
\text{WD}=\int_{-\infty}^{+\infty}|P-Q|
\end{aligned}
\end{equation}
for distributions $p$ and $q$ with cumulative distribution functions (CDF) $P$ and $Q$, respectively. The continuous form of WD can also be adapted for discrete cases.

\subsection{Implementation}

The graph encoder is built with PyTorch Geometric \cite{fey2019fast} modules. The encoder is composed of three GAT/UniMP layers, each followed by batch normalization \cite{ioffe2015batch} and a non-linear activation \cite{he2015delving}. The sequence decoder is a stack of six self-attention layers, where each layer has subsequent layer normalization and feed-forward layers. The whole model is trained with the Adam optimizer \cite{kingma2014adam}, on Nvidia V100 GPU. 

\subsection{Prediction Results}

\begin{table}[t]
\centering
\begin{minipage}{.95\linewidth}
\centering
\scalebox{0.96}{
\begin{tabular}{l|llll}
\hline\hline
model & $R^2$ $\uparrow$ & MAE $\downarrow$ & MSE $\downarrow$ & WD $\downarrow$ \\ 
\hline
E3NN & 0.560 & 0.00869 & 0.000389 & 0.1217 \\
E3NN-new & 0.621 & 0.00833 & 0.000275 & 0.0765 \\
GATGNN & 0.457 & 0.00976 & 0.000405 & 0.1376 \\
E3NN2DoS & \underline{0.660} & \underline{0.00740} & \underline{0.000247} & 0.0740 \\ 
Mat2Spec & 0.641 & 0.00773 & 0.000264 & \underline{0.0736} \\
Xtal2DoS & \textbf{0.734} & \textbf{0.00663} & \textbf{0.000193} & \textbf{0.0631} \\ \hline
improve & 11.21\% & 10.41\% & 21.86\% & 14.27\% \\ \bottomrule
\end{tabular}
}
\caption{The evaluation results of four metrics on phDoS.
(Best in bold; 2nd best underlined.)}
\label{tab:phdos_eval}
\end{minipage}
~\\
\begin{minipage}{.95\linewidth}
\centering
\scalebox{0.96}{
\begin{tabular}{l|llll}
\hline\hline
model & $R^2$ $\uparrow$ & MAE $\downarrow$ & MSE $\downarrow$ & WD $\downarrow$\\ \hline
E3NN & 0.410 & 4.8968 & 99.852 & 0.467 \\
E3NN-new & \underline{0.555} & 3.9190 & \underline{74.949} & 0.233 \\
GATGNN & 0.492 & 4.2965 & 85.555 & 0.285 \\
E3NN2DoS & 0.552 & \underline{3.8900} & 75.453 & 0.236 \\ 
Mat2Spec & 0.545 & 3.8939 & 76.624 & \underline{0.230} \\
Xtal2DoS & \textbf{0.577} & \textbf{3.7845} & \textbf{71.213} & \textbf{0.218} \\ \hline
improve & 3.96\% & 2.71\% & 4.98\% & 5.22\% \\ 
\bottomrule
\end{tabular}
}
\caption{The evaluation results of four metrics on eDoS.
(Best in bold; 2nd best underlined.)}
\label{tab:edos_eval}
\end{minipage}
\end{table}


Xtal2DoS is first evaluated on phDoS, compared with other state-of-the-art baselines, including E3NN, GATGNN, Mat2Spec and some of their variants. Among all the compared methods, Xtal2DoS gives the best numbers. On $R^2$, MAE, MSE and WD, the relative improvements over the best compared baseline are as large as 11.21\%, 10.41\%, 21.86\%, 14.27\%, respectively. The average improvement is 14.44\%.

Table \ref{tab:edos_eval} shows the performance gain achieved by Xtal2DoS over other competitive baselines on the more challenging eDoS dataset. Xtal2DoS outperformed the best baseline on the four metrics by 3.96\%, 2.71\%, 4.98\%, 5.22\% respectively. The average improvement is 4.22\%, which is a noticeable increment especially when the training speed is faster (see section \ref{sec:speed}).

\subsection{Ablation Study}

\begin{table}[t]
\centering
\begin{minipage}{.95\linewidth}
\centering
\scalebox{0.95}{
\begin{tabular}{l|cccc}
\hline\hline
model & $R^2$ $\uparrow$ & MAE $\downarrow$ & MSE $\downarrow$ & WD $\downarrow$\\ 
\hline
RNN & 0.663 & 0.00794 & 0.000245 & 0.0797 \\
RNN+Attn & 0.668 & 0.00726 & 0.000241 & 0.0664 \\
Chunk RNN+Attn & \underline{0.693} & \underline{0.00720} & \underline{0.000223} & \underline{0.0656} \\ 
Xtal2DoS & \textbf{0.734} & \textbf{0.00663} & \textbf{0.000193} & \textbf{0.0631} \\ \hline
improve & 5.92\% & 7.92\% & 13.45\% & 3.81\% \\ \hline
\end{tabular}
}
\caption{Ablation study on phDoS.}
\label{tab:phdos_abla}
\end{minipage}
~\\
\begin{minipage}{.95\linewidth}
\centering
\scalebox{0.95}{
\centering
\begin{tabular}{l|cccc}
\hline\hline
model & $R^2$ $\uparrow$ & MAE $\downarrow$ & MSE $\downarrow$ & WD $\downarrow$\\ 
\hline
Chunk RNN+Attn & 0.558 & 3.9125 & 75.013 & 0.221 \\
Xtal2DoS & \textbf{0.577} & \textbf{3.7845} & \textbf{71.213} & \textbf{0.218} \\ \hline
improve & 3.41\% & 3.27\% & 5.07\% & 1.36\% \\ \hline
\end{tabular}
}
\caption{Ablation study on eDoS.}
\label{tab:edos_abla}
\end{minipage}
\end{table}

\textbf{Encoder} ~ To validate the neural module selection of Xtal2DoS, we compare E3NN graph encoder (E3NN2DoS) and UniMP (Xtal2DoS) graph encoder with transformer decoder fixed. E3NN2DoS adopts the new version of E3NN. From Table \ref{tab:phdos_eval} and \ref{tab:edos_eval}, Xtal2DoS consistently outperforms E3NN2DoS. Though E3NN preserves the extra information like angles, these features might also lead to overfitting. E3NN could potentially work better when the data are more scarce and the properties of interest are scalars. Similar findings can also be found in \cite{yan2022periodic}.

\textbf{Decoder} ~ We further compare the performance of RNN, RNN+attention, Chunk RNN+attention, and the transformer (Xtal2DoS). We use the same four metrics to compare these settings. In Table \ref{tab:phdos_abla} and \ref{tab:edos_abla}, transformers always dominate other RNN-based models on all the metrics. The improvements shown in the last row indicate the comparison between Xtal2DoS and the best compared model (usually Chunk RNN+Attn).

\subsection{Qualitative Results}

We also show some qualitative results on both phDoS and eDoS in Figure \ref{fig:phdos_fit} and \ref{fig:edos_fit} (in Appendix). Both figures demonstrate the accuracy of our predictions. The red and blue lines are always closely correlated. 

\subsection{Training Speed}
\label{sec:speed}

\begin{table}[t]
\centering
\scalebox{0.95}{
\begin{tabular}{c|cccccc}
\hline
model & E3NN & Mat2Spec & RNN & RNN+Attn & Chunk RNN+Attn & \textbf{Xtal2DoS} \\ \hline
Time per epoch (s) & 31 & 33 & 54 & 69 & 42 & \textbf{11} \\\hline
\end{tabular}
}
\caption[Training time for RNN-based and transformer-based models]{The training time for Mat2Spec, RNN-based sequence models and Xtal2DoS, on phDoS. For fair comparison, we illustrate the training cost for each epoch in seconds. They take a similar number of epochs to converge. So, overall, Xtal2DoS is the fastest method among all the  models.}
\label{tab:train_time_xtal2dos}
\end{table}

One major advantage of the transformer models is parallelism. Unlike RNNs where each step has to wait for the previous output, the transformer computes the outputs at all the steps simultaneously and enforces  auto-regressiveness through masking. These operations can be performed in matrix form, taking advantage of powerful GPUs, and thus substantially reduce the training cost. Table \ref{tab:train_time_xtal2dos} shows the training time per epoch on phDoS for Mat2Spec, RNN-based sequence models and Xtal2DoS. Mat2Spec does not involve any sequential model and generates outputs through MLP, so it is faster than any RNN-based sequence decoder model. Mat2Spec instead integrates  probabilistic sampling and contrastive learning, so it is not surprising that it takes more time than Xtal2DoS. Unsurprisingly, RNN trains more than 50\% slower than Mat2Spec. Adding the attention module further prolongs the time cost. RNN+Attention doubles the training cost compared with Mat2Spec. This also motivated us to attempt chunk RNN. As a result, chunk RNN+Attention can complete a training epoch faster than RNN and RNN+Attention, but is still slower than Mat2Spec. It only takes one third of Mat2Spec's training time, one fifth of RNN's training time, and one fourth of chunk RNN+Attention's training time. Xtal2DoS excels at speed because of its parallel design and its succinct structure to avoid extra neural module. Faster models are often more advantageous in training, tuning and deploying.


\section{Conclusion}

In this work, we introduce a novel graph-to-sequence learning framework, \textit{Xtal2DoS}, for predicting the spectral properties of materials, \textit{phDoS} and \textit{eDoS}, from the input crystalline structures. Our model adopts an improved graph attention network (to handle edges) as the graph encoder, and a transformer as the decoder for sequential decoding. Xtal2DoS is fully attentional and captures all the atom-atom, atom-sequence and step-step correlations. Therefore, it considerably exceeds other existing state-of-the-art baselines, in quantitative evaluation metrics, qualitative matching results and training speed.

\bibliographystyle{unsrtnat}
\bibliography{ref.bib}


\clearpage
\appendix

\section{Qualitative Results}

\begin{figure}[ht]
  \centering
  \includegraphics[width=0.85\linewidth]{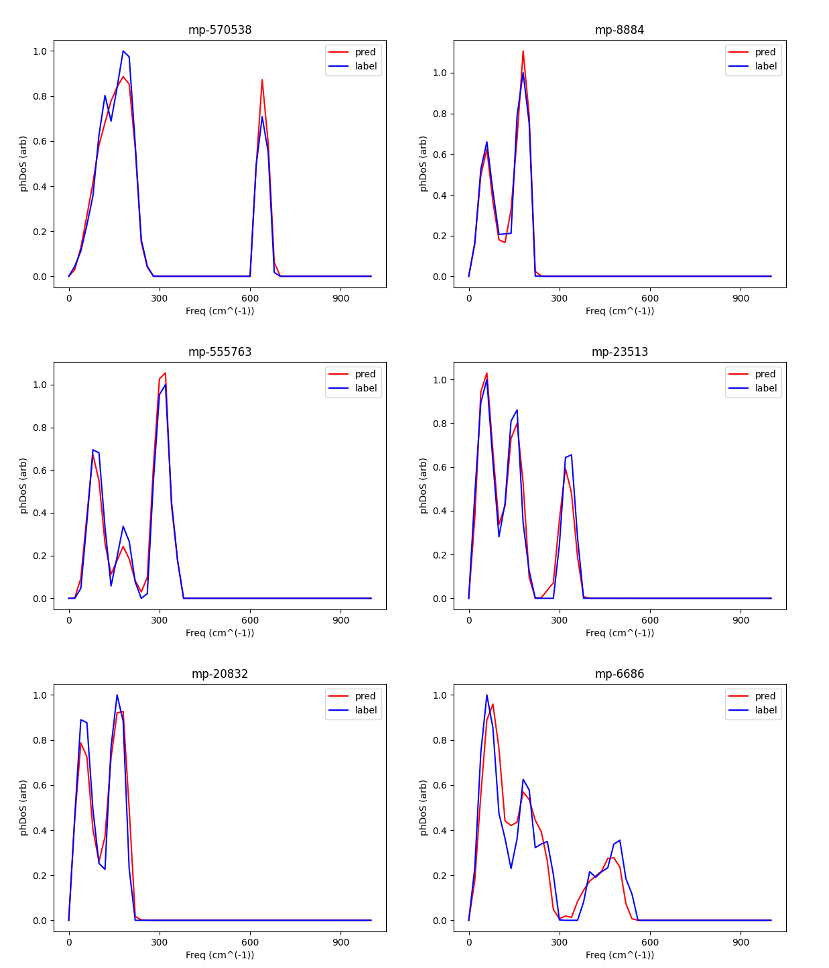}
  \caption[Qualitative results on phDoS]{Comparison of the phDoS predictions and ground-truth labels on some crystalline systems. The red line denotes the prediction and the blue line denotes the ground-truth.}
  \label{fig:phdos_fit}
\end{figure}

\begin{figure}[ht]
  \centering
  \includegraphics[width=0.85\linewidth]{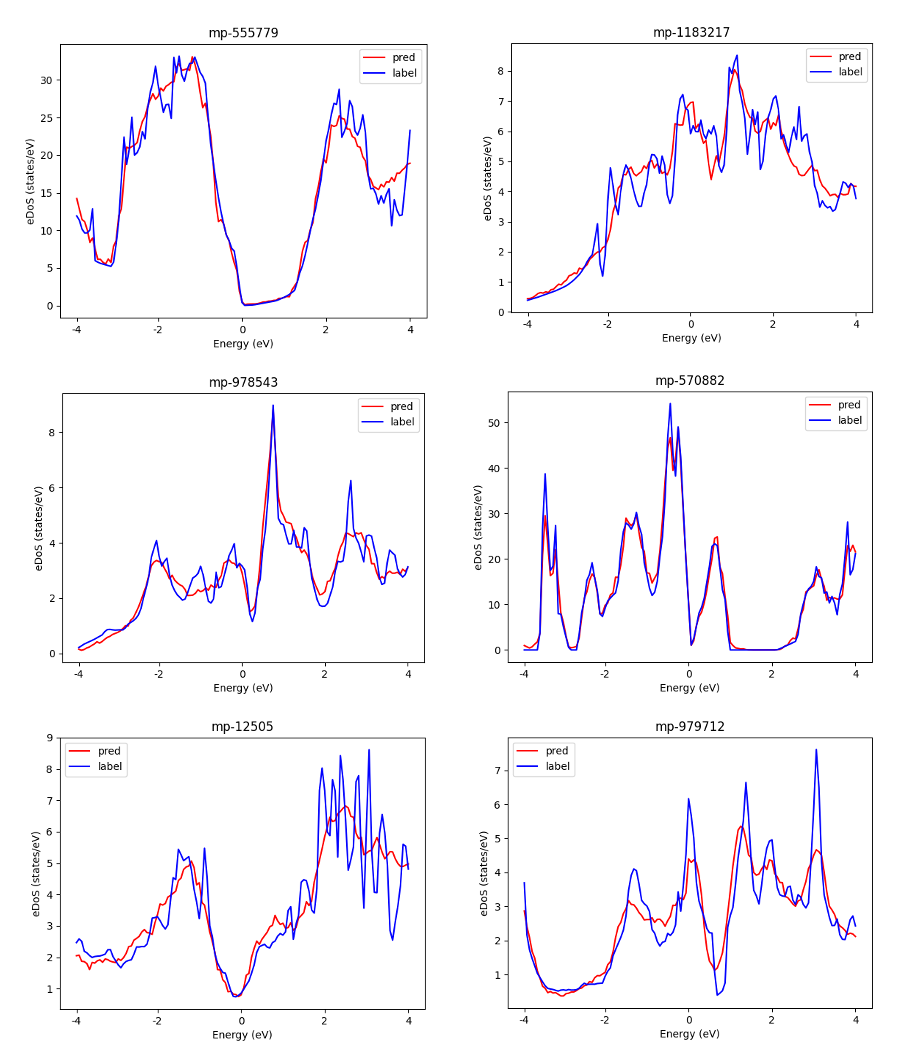}
  \caption[Qualitative results on eDoS]{The comparison between the eDoS predictions and ground-truth labels on some crystalline systems. The red line denotes the prediction and the blue line denotes the ground-truth.}
  \label{fig:edos_fit}
\end{figure}

\end{document}